\documentclass[11pt]{article}

\usepackage[final]{acl}

 \usepackage{microtype}
 \usepackage{subcaption}

\usepackage[english,bidi=default]{babel} 
\babelfont{rm}{TeXGyreTermesX} 
\babelprovide[import]{hindi}
\babelfont[*devanagari]{rm}{Lohit Devanagari}
\babelprovide[import]{arabic}
\babelfont[*arabic]{rm}{Noto Sans Arabic}
\babelprovide[import]{chinese}
\babelfont[*chinese]{rm}{Noto Serif CJK TC}

\newfontfamily\ipafont{NotoSerif-Regular.ttf}[
    Path = ./fonts/]

\usepackage{inconsolata}
\usepackage[dvipsnames]{xcolor}
\usepackage{url}
\usepackage{graphicx}
\usepackage{multirow}

\usepackage{booktabs}
\usepackage{amsmath}

\usepackage{pgf}
\usepackage{colortbl}

\newcommand{\gradientcell}[6]{%
    \def\value{#1}%
    \def\minvalue{#2}%
    \def\maxvalue{#3}%
    \def\mincolor{#4}%
    \def\maxcolor{#5}%
    \def\transparency{#6}%
    \ifdimcomp{\value pt}{>}{\maxvalue pt}{\cellcolor{#5!100.0!#4!#6}\value}{%
    \ifdimcomp{\value pt}{<}{\minvalue pt}{\cellcolor{#5!0.0!#4!#6}\value}{%
         \pgfmathparse{int(round(100*(#1/(\maxvalue-\minvalue))-(\minvalue *(100/(\maxvalue-\minvalue)))))}%
        \xdef\tempa{\pgfmathresult}%
        \cellcolor{#5!\tempa!#4!#6}\value%
    }}%
}

\definecolor{LightGray}{gray}{0.95} 
\newcommand{\negcorr}[1]{\gradientcell{#1}{-1.0}{0.0}{MidnightBlue}{LightGray}{50}}
\newcommand{\poscorr}[1]{\gradientcell{#1}{0.}{1.0}{LightGray}{BrickRed}{50}}
\newcommand{\C}[1]{%
  \ifdim#1pt<0pt
    \negcorr{#1}%
  \else
    \poscorr{#1}%
  \fi
}

\usepackage{tikz}

\newrobustcmd*{\square}[1]{\tikz{\filldraw[draw=#1,fill=#1] (0,0)
rectangle (0.2cm,0.2cm);}}

\newrobustcmd*{\mycircle}[1]{\tikz{\filldraw[draw=#1,fill=#1] (0,0) circle [radius=0.1cm];}}

\newrobustcmd*{\mytriangle}[1]{\tikz{\filldraw[draw=#1,fill=#1] (0,0) --
(0.2cm,0) -- (0.1cm,0.2cm);}}

\newrobustcmd*{\mydiamond}[1]{%
  \tikz{\filldraw[draw=#1,fill=#1] 
    (0.1cm,0) -- (0.2cm,0.1cm) -- (0.1cm,0.2cm) -- (0cm,0.1cm) -- cycle;}}
    
\title{Automated Quality Control for Language Documentation: Detecting Phonotactic Inconsistencies in a Kokborok Wordlist}

\author{
  Kellen Parker van Dam\textsuperscript{1} \and
  Abishek Stephen\textsuperscript{2} \\
  \textsuperscript{1}Chair for Multilingual Computational Linguistics, University of Passau, Germany \\
  \textsuperscript{2}Institute of Formal and Applied Linguistics, Charles University, Czech Republic \\
}

\begin{document}

\maketitle
\begin{abstract}
Lexical data collection in language documentation often contains transcription errors and  borrowings that can mislead linguistic analysis. We present unsupervised methods to identify phonotactic inconsistencies in wordlists, applying them to a multilingual dataset of Kokborok varieties with Bangla. Using phoneme-level and syllable-level n-gram language models, our approach identifies potential transcription errors and borrowings. We evaluate our methods using hand annotated gold standard and rank the phonotactic outliers using precision and recall at K metric. The ranking approach provides field linguists with a method to flag entries requiring verification, supporting data quality improvement in low-resourced language documentation.
\end{abstract}

\section{Introduction}

In linguistic fieldwork, description frequently begins with the collection of lexical data \citep{chelliah2014fieldwork}. This is often done by means of concept lists such as those of Swadesh \citep{swadesh1955towards}. In the early stages of research, initial elicitation sessions commonly produce “messy” data. When lexical items are collected as a preliminary step, the fieldworker may not be fully acquainted with the phonological system of the target language. Which sounds are phonemic as opposed to allophonic variation may be uncertain. As a result, words are often transcribed narrowly which may be inconsistent from entry to entry regarding the underlying phonemes, as well as potentially failing to capture the underlying phonemic contrasts. A lack of systematicity in the transcription can in turn create issues later on in data analysis \citep{himmelmann1998documentary}.  

For proper documentation it is important to incorporate the speech of multiple participants. However, when data is drawn from multiple speakers in this manner, differences in dialect or accent may be reflected in the forms recorded. Variation between careful and casual speech styles can introduce further irregularities \cite{chelliah2014fieldwork}. 


Lexical borrowing constitutes another potential complication. Borrowed terms may have varying degrees of adherence to the underlying phonemic system. Terms may also have been borrowed twice, with an intermediate borrowing of another closely related language having a different set of phonotactic constraints, thus obscuring their borrowed nature. Thus, it is important that borrowings can be readily identified when attempting to understand the phonology of a language. Borrowed forms may enter the dataset without the researcher’s awareness, particularly when the donor language is unfamiliar to the fieldworker. 


The context of elicitation is also important. Differences in approaches can exert a significant influence on the quality and consistency of the data. The degree of formality in the interaction, the presence of other speakers, and the level of fatigue or attention on the part of the consultant can all affect the data. The fieldworker’s own background and expectations also shape the data in subtle but consequential ways \citep{KellyLahaussois2021}.



For these reasons, having a method for detection of phonological outliers is of great value to the documentary linguist. By identifying potential borrowings or inconsistencies introduced by factors, the end result of any descriptive study is immediately aided in the very first steps of lexical data collection. Having automated flags for ``this entry looks phonotactically weird'' could save field linguists considerable time, especially when working with under-resourced languages where you can't rely on external data verification. We do this detection using n-gram language modeling based on phoneme and syllable-level analysis.

\section{Related Work}
The current research deals with identifying the phonotactic inconsistencies in a linguistic wordlist which in a different light can be seen to have concordances with spelling checkers or borrowing detection methods. Our work however, is not aimed towards either of them albeit the overt similarities. Worth mentioning are some attempts of borrowing detection using wordlists. \citet{miller2021neural} where automatic methods for detecting lexical borrowings from monolingual wordlists, comparing different neural network based architectures.  \citet{list2019automated} presents approaches for detecting language contact and borrowing, focusing on phylogenetic networks, sequence comparison methods for detecting borrowings in multilingual wordlists, and trait-based approaches that distinguish borrowed from inherited features using borrowability arguments.  

\section{Source Data}
\label{sec:3}
We rely on data for Kokborok (Glottocode: \texttt{\href{https://glottolog.org/resource/languoid/id/tipp1238}{tipp1238}}, \citealp{hammarstrom2025glottolog}), an under-described Tibeto-Burman language group  under the Barish language branch \citep{delanceybarish}.

The consonant inventory is relatively moderate in size, with a notable series of aspirated stops that likely developed through Indo-Aryan influence, as aspiration contrasts are less common in many Tibeto-Burman languages. The language maintains voicing distinctions across bilabial, dental, velar, and palatal places of articulation. Word-finally, however, obstruents typically devoice, a pattern not found in neighboring Bangla. Notably, voiced affricates like \textit{\ipafont{/dʒ/}} are not native to Kokborok but appear in Bangla loanwords, representing sounds borrowed along with vocabulary. Kokborok strongly prefers open syllables and avoids consonant clusters, reflecting its Tibeto-Burman phonotactic constraints. This contrasts sharply with Bangla, which permits complex consonant clusters both word-initially and word-finally. Where Bangla allows syllables like \textit{\ipafont{/bdʒro/}} or final clusters like \textit{\ipafont{/-sto/}}, Kokborok maintains simpler CV(C) structures with very limited coda positions.

These phonotactic differences create a clear phonological boundary between Kokborok and Bangla despite their geographic proximity. The result is two typologically distinct systems coexisting in close contact, with Kokborok maintaining its characteristic Tibeto-Burman simplicity in syllable structure even while absorbing lexical material from its Indo-Aryan neighbor.

Our data comes from \citealp{Zenodo_17973868}, a sociolinguistic survey of 306 concepts in 20 Kokborok varieties plus 3 varieties of Garo (\texttt{\href{https://glottolog.org/resource/languoid/id/garo1247}{garo1247}}), and standard Bangla as the main contact language. This concept list is a good representation of the language as it covers the majority of basic morphemes which go into lexical construction. As is typical in Tibeto-Burman languages of the region, words are primarily compounds of simpler common morphemes. This set of 306 concepts is considerably larger than the amount that would normally go into computational phylogenetic work, such as the 180 concepts of \citet{sagart2019dated} or the 100 of \citet{galucio2015genealogical}, and coverts the full range of phonological variation occurring in native forms. 

Kokborok data were converted to the Cross-Linguistic Data Format (CLDF; \citet{Forkel2018CLDF}) by the authors. Morphological features external to the citation form were removed, as were erroneous repeated diacritics which did not contribute to the transcription. The data were otherwise not modified, leaving ambiguous transcriptions\footnote{For example in the AbengGajni doculect, the verb `to eat' is erroneously transcribed as \textit{\ipafont{tshaʔ}} with no clear indication if this should be \textit{\ipafont{tʃaʔ}} or \ipafont{\textit{ts\textsuperscript{h}aʔ}}.} as is. The CLDF dataset is available in a GitHub repository\footnote{\url{https://github.com/phonemica/kimkokborok}. The dataset can also be accessed here-\url{https://doi.org/10.5281/zenodo.17973867}.}







\section{Experiments} 



The identification of the phonological anomalies or outliers in our dataset proceeds via implementing simple n-gram language models at the phoneme and syllable levels. First, we run the experiments on the phoneme level which aims to capture rare phoneme sequences and then we contrast it with positional phonotactics using the syllable-level analysis that captures more linguistically motivated violations. The source codes are publicly available\footnote{\url{https://github.com/abishekjs/kokborok-anotect}}.

The training data has 3055 words after removing duplicates\footnote{Since the linguistic varieties or doculects are closely related, the same word forms are used to encode a given semantic concept.}. The gold data contains 555 words marked as borrowings. Our annotations focus exclusively on borrowings, as these were straightforward to identify given the clear phonological and lexical distinctions between Kokborok and Bangla, the primary source of loanwords in the dataset. We treat words in Kokborok that are almost identical to the Bangla counterpart phonologically (as explained in §~\ref{sec:3}) for a given semantic concept as borrowed. For example, the word for `rainbow' in Bangla and doculect MukchakBarbakpur is \textit{\ipafont{ɾɔŋdʰɔnu}} and hence marked as a borrowing. Transcription errors were not systematically annotated due to the difficulty of distinguishing genuine errors from dialectal variation or unknown phonological processes, making borrowings more reliable for evaluating our methods. 

\subsection{Phoneme-level N-gram Language Modeling}
To identify phonotactic anomalies such as transcription errors and borrowings, we train phoneme bigram and trigram language models on the Kokborok data. Words are padded with boundary markers (e.g.,\textit{ \ipafont{\^{}noukʰa\$}}), and diacritics are treated as separate characters. We apply Laplace smoothing to handle unseen n-grams. 

Our mathematical assumption is that the words with transcription errors or borrowings would have some phoneme sequences which are rare in the language, and using the negative log likelihood (NLL) such words would be flagged when ranking by resulting NLL scores. This can help linguists make quick quality checks, as the stronger outliers would be captured in the top K words. We compute NLL for words using different aggregation methods to capture character-level variations.

\paragraph{Arithmetic Mean} captures the expected information content per n-gram. It normalizes for word length enabling fair comparisons between long and short words. 

\begin{equation}
     \text{Mean NLL} = -\frac{1}{N}\sum_{i=1}^{N} \log_2 p(x_i)
\end{equation}
where $N$ is the number of n-grams in the word and $x_i$ represents the $i$-th n-gram.

\paragraph{Harmonic Mean} emphasizes the most typical n-grams within a word, being heavily weighted toward smaller NLL values. This metric is particularly useful for identifying words that contain a core of native phonotactic patterns even when some unusual or rare n-grams are present.

\paragraph{Min} identifies the most typical n-gram in a word, revealing whether the word shares any common phonotactic patterns with the native vocabulary. Even heavily borrowed words may contain some typical n-grams, and this metric helps assess the degree of phonotactic integration of loanwords into the native system. A very low minimum NLL suggests the word has at least partial structural overlap with native phonotactics.

\paragraph{Max} identifies the most atypical n-gram in a word, a rare n-gram can be a reminiscent of the source language in case of the word being borrowed. It could also potentially flag off partially integrated loanwords.  

\subsection{Syllable-level N-gram Language Modeling}
We implement automatic syllabification based on the sonority hierarchy and maximum onset principle, where syllable boundaries are determined by identifying sonority peaks and applying language-universal syllable structure constraints. The syllable boundary symbol (\texttt{.}) serves as a structural marker. Here too, we add boundary markers and use the aggregation methods used in the phoneme level analysis.

\subsubsection{Analysis Types}
We employ three distinct approaches to analyze phonotactic patterns:

\begin{itemize}
    \item \textbf{Within-syllable analysis}: We calculate negative log likelihood of character n-grams that occur strictly within individual syllables, respecting syllable boundaries and focusing on internal syllabic structure.
    
    \item \textbf{Cross-boundary analysis}: We extract character n-grams that span across syllable boundaries, capturing phonotactic patterns that violate typical syllable constraints and may indicate borrowing or transcription anomalies.
    
    \item \textbf{Boundary-as-phoneme analysis}: We treat syllable boundaries as legitimate phonemes in the sequence, allowing n-grams to include the syllable boundary symbol and capturing positional sensitivity at syllable edges.
\end{itemize}

\begin{table*}[ht]
\centering

\caption{Precision and Recall at K for different NLL aggregation methods and baselines for the phoneme-level n-gram models.}
\footnotesize
\label{tab:results_phoneme}
\setlength{\tabcolsep}{4.8pt}
\begin{tabular}{@{}llcccccc@{}}
\toprule
\textbf{N-gram} & \textbf{Method} & \textbf{P@100} & \textbf{P@500} & \textbf{P@1000} & \textbf{R@100} & \textbf{R@500} & \textbf{R@1000} \\
\midrule
\multirow{6}{*}{Bigram} 
& Arithmetic Mean & {\C{0.43}} & {\C{0.32}} & {\C{0.26}} & {\C{0.08}} & {\C{0.29}} & {\C{0.47}} \\
& Harmonic Mean & {\C{0.43}} & \C{0.31} & \C{0.26} & {\C{0.08}} & \C{0.28} & \C{0.47} \\
& Min NLL & \C{0.36} & {\C{0.32}} & \C{0.23} & \C{0.06} & {\C{0.29}} & \C{0.42} \\
& Max NLL & \C{0.32} & \C{0.26} & \C{0.23} & \C{0.06} & \C{0.24} & \C{0.42} \\
\cmidrule(lr){2-8}
& Uniform Random & \C{0.15} & \C{0.19} & \C{0.17} & \C{0.03} & \C{0.17} & \C{0.31} \\
& Stratified Random & \C{0.17} & \C{0.20} & \C{0.17} & \C{0.03} & \C{0.18} & \C{0.30} \\
\midrule
\multirow{6}{*}{Trigram} 
& Arithmetic Mean & \C{0.45} & {\C{0.33}} & {\C{0.28}} & \C{0.08} & {\C{0.30}} & {\C{0.51}} \\
& Harmonic Mean & {\C{0.46}} & \C{0.32} & {\C{0.29}} & {\C{0.08}} & \C{0.28} & {\C{0.52}} \\
& Min NLL & \C{0.40} & \C{0.32} & \C{0.27} & \C{0.07} & \C{0.29} & \C{0.49} \\
& Max NLL & \C{0.20} & \C{0.29} & \C{0.25} & \C{0.04} & \C{0.26} & \C{0.46} \\
\cmidrule(lr){2-8}
& Uniform Random & \C{0.20} & \C{0.17} & \C{0.19} & \C{0.04} & \C{0.15} & \C{0.34} \\
& Stratified Random & \C{0.32} & \C{0.21} & \C{0.19} & \C{0.06} & \C{0.19} & \C{0.34} \\
\bottomrule
\end{tabular}
\end{table*}

\subsection{Results}
For the field linguists, it would be highly efficient to discover anomalies based on the ranking of the words following their NLL scores observed for all of the aggregation methods. To facilitate that we use recall and precision at K as our evaluation metric. We use the mean NLL as the baseline for the experiments. The gold data is hand annotated, the words borrowed from Bangla are labeled as borrowings. The current dataset do not have any transcription errors, but the assumption and also the strong caveat of our method also ensures the flagging of such errors. 

We establish two random sampling baselines to evaluate whether our n-gram phonotactic models perform better than chance. The uniform random baseline samples K words randomly from the wordlist without any prior assumptions, supporting the hypothesis where all words are equally likely to be anomalies. The stratified random baseline samples words proportionally by length, controlling for the possibility that transcription errors or borrowings may be biased toward longer or shorter words. 

\subsection{Phoneme-level Results}
Table~\ref{tab:results_phoneme} demonstrates that our phoneme-level n-gram phonotactic models substantially outperform random baselines in identifying phonotactic anomalies. Trigram models achieve the strongest performance, with precision at 100 reaching 0.46 and recall at 1000 reaching 0.52 using harmonic mean aggregation. The superiority of trigrams over bigrams suggests that richer phonotactic context is crucial for capturing constraints violations, while the consistent performance of arithmetic and harmonic mean aggregation indicates that anomalies are characterized by sustained phonotactic unusualness across the entire word rather than isolated rare n-grams. At K=500, our best model achieves 33\% precision, meaning that approximately one in three flagged items is a genuine anomaly. However, the plateau at 52\% recall suggests that roughly half of the gold anomalies are phonotactically well-formed, indicating they may represent semantic borrowings or transcription errors that do not violate native phonological constraints.

The best performing model based on trigram-harmonic mean identifies words like \textit{\ipafont{ʤʰaɽu}}, \textit{ \ipafont {oʃud}}, \textit{ \ipafont{ʈɪkʈɪki}}, \textit{ \ipafont{ɔnɛk}}, \textit{ \ipafont{mɛgʰgɔɹʤon}} and so on in the top 100 words being flagged as anomalous. In the top 500 words like \textit{  \ipafont {poɾibar}}, \textit{  \ipafont {murgi}}, \textit{  \ipafont {gɔɾom}} get flagged. This ranking pattern reflects the model's sensitivity to different degrees of phonotactic deviations, top 100 of the flagged words typically contain phonemes or phoneme combinations that are extremely rare or absent in native Kokborok vocabulary (such as retroflex consonants and aspirated affricates), while words ranked in the top 500 exhibit more subtle violations involving less frequent but attested phoneme sequences, suggesting partial phonological adaptation common for borrowings.

\begin{table*}[ht]
\centering
\caption{Precision and Recall at K for bigram models with arithmetic mean aggregation across different syllable analysis types.}
\footnotesize
\label{tab:results_syllable_bigram}
\begin{tabular}{@{}lcccccc@{}}
\toprule
\textbf{Analysis} & \textbf{P@100} & \textbf{P@500} & \textbf{P@1000} & \textbf{R@100} & \textbf{R@500} & \textbf{R@1000} \\
\midrule
Within & {\C{0.47}} & {\C{0.32}} & {\C{0.26}} & {\C{0.08}} & {\C{0.29}} & {\C{0.47}} \\
Cross & \C{0.21} & \C{0.20} & \C{0.18} & \C{0.04} & \C{0.18} & \C{0.33} \\
Boundary & {\C{0.50}} & \C{0.29} & \C{0.24} & {\C{0.09}} & \C{0.26} & \C{0.43} \\
\midrule
Uniform Random & \C{0.15} & \C{0.19} & \C{0.17} & \C{0.03} & \C{0.17} & \C{0.31} \\
Stratified Random & \C{0.17} & \C{0.20} & \C{0.17} & \C{0.03} & \C{0.18} & \C{0.30} \\
\bottomrule
\end{tabular}
\end{table*}

\subsection{Syllable-level Results}
Our results (Table~\ref{tab:results_syllable}, see Appendix) reveal distinct performance patterns across three phonotactic modeling approaches. Within-word analysis achieves the strongest overall performance with precision at 100 of 0.47 for bigrams and recall at 1000 of 0.49 for trigrams, effectively capturing internal phonological structure violations. Boundary-as-phoneme analysis shows competitive results, particularly at lower K values where trigram models reach precision at 100 of 0.49, indicating that syllable boundary constraint violations are highly predictive of anomalies. In contrast, cross-boundary analysis substantially underperforms, with precision rarely exceeding 0.30 and recall at 1000 capped at 0.42, suggesting that phonotactic violations are better characterized by position-specific patterns within syllable constituents rather than by boundary-crossing transitions alone.

Words like \textit{ \ipafont{mɛgʰ}} and \textit{ \ipafont {moɾɪʧ}} are caught early on at top 100 using the boundary-as-phoneme bigram arithmetic mean setup (Table~\ref{tab:results_syllable_bigram}). These words were flagged off in the top 500 of the phoneme trigram harmonic mean setup. This demonstrates the complementary strengths of syllable-aware modeling. The boundary marker (.) itself being present in some sequence is highly influential on results in that it closely reflects syllable position i.e. the phonemes preceding (.) indicate syllable onset or nucleus, following (.) indicates coda position and so on. 

\section{Conclusion}
Our study addresses transcription errors and unidentified borrowings that can skew typological analysis in wordlist-based language documentation. Designed for the initial stages of data collection, these methods provide field linguists with systematic tools to identify entries requiring closer inspection. Our results reveal that phoneme-level n-gram models capture most anomalies also flagged by syllable-level models, suggesting that while incorporating explicit phonotactic knowledge through syllabification provides some benefit, raw phoneme sequence modeling alone achieves comparable performance. This indicates that computationally simpler approaches may be sufficient for practical anomaly detection in fieldwork settings, though the syllable-level analysis offers additional interpretability by identifying specific constraint violations. 

\section*{Limitations}
Due to the limited size of documentary wordlists, data-intensive approaches such as neural language models cannot be applied, though such methods might yield superior performance in high-resource settings. Borrowing detection is inherently challenging making gold standard creation labor-intensive. However, the statistical nature of n-gram models ensures they capture phonotactic inconsistencies providing field linguists with a tool for identifying entries that deviate from expected patterns and require closer inspection.

In terms of the method's usefulness in borrowing detection, this relies heavily on having a phonotactically more restrictive language borrowing from one with a greater possibility of sounds, or simply sounds which are not found in the borrowing language. Detecting Kra-Dai borrowings into a phonologically similar Tibeto-Burman language would not be possible in this case. However, the method is still useful in detecting transcription errors, novel sound changes, dialectal variation, or other cases which still warrant further investigation by the linguist even if not the result of borrowing.

Finally, as the fieldworker’s expectations also shape how they transcribe data \citep{KellyLahaussois2021}, application of this approach too quickly or without further investigation into the language could result in further over-application of mistakes. One should not blindly assume anomalies are errors. Rather, they are points to be investigated further and confirmed.

\section*{Acknowledgments}
We thank the anonymous reviewers for their valuable feedback.
This research was supported by the Charles University student project GA UK No.\ 101924 and partially supported by SVV project number 260 698. 
\bibliography{custom}

\appendix

\section{Appendix}
\label{sec:appendix}

\begin{table*}[ht]
\centering
\caption{Precision and Recall at K for different analysis types and n-gram sizes for the syllable-level n-gram models.}
\footnotesize
\label{tab:results_syllable}
\setlength{\tabcolsep}{3.5pt}
\begin{tabular}{@{}lllcccccc@{}}
\toprule
\textbf{Analysis} & \textbf{N-gram} & \textbf{Method} & \textbf{P@100} & \textbf{P@500} & \textbf{P@1000} & \textbf{R@100} & \textbf{R@500} & \textbf{R@1000} \\
\midrule
\multirow{8}{*}{Within} 
& \multirow{4}{*}{Bigram} 
& Arithmetic Mean & {\C{0.47}} & {\C{0.32}} & {\C{0.26}} & {\C{0.08}} & {\C{0.29}} & {\C{0.47}} \\
&& Harmonic Mean & {\C{0.43}} & {\C{0.32}} & {\C{0.26}} & {\C{0.08}} & {\C{0.29}} & {\C{0.48}} \\
&& Min NLL & \C{0.35} & {\C{0.32}} & \C{0.23} & \C{0.06} & {\C{0.29}} & \C{0.42} \\
&& Max NLL & \C{0.35} & \C{0.28} & \C{0.23} & \C{0.06} & \C{0.26} & \C{0.41} \\
\cmidrule(lr){2-9}
& \multirow{4}{*}{Trigram} 
& Arithmetic Mean & {\C{0.45}} & {\C{0.31}} & {\C{0.27}} & {\C{0.08}} & {\C{0.28}} & {\C{0.49}} \\
&& Harmonic Mean & {\C{0.45}} & {\C{0.31}} & {\C{0.27}} & {\C{0.08}} & {\C{0.28}} & {\C{0.48}} \\
&& Min NLL & \C{0.40} & {\C{0.32}} & {\C{0.27}} & \C{0.07} & {\C{0.29}} & {\C{0.49}} \\
&& Max NLL & \C{0.37} & \C{0.29} & \C{0.23} & \C{0.07} & \C{0.26} & \C{0.42} \\
\midrule
\multirow{8}{*}{Cross} 
& \multirow{4}{*}{Bigram} 
& Arithmetic Mean & \C{0.21} & \C{0.20} & \C{0.18} & \C{0.04} & \C{0.18} & \C{0.33} \\
&& Harmonic Mean & \C{0.21} & \C{0.20} & \C{0.19} & \C{0.04} & \C{0.18} & \C{0.33} \\
&& Min NLL & \C{0.22} & \C{0.19} & \C{0.17} & \C{0.04} & \C{0.17} & \C{0.30} \\
&& Max NLL & \C{0.34} & \C{0.21} & \C{0.19} & \C{0.06} & \C{0.19} & \C{0.34} \\
\cmidrule(lr){2-9}
& \multirow{4}{*}{Trigram} 
& Arithmetic Mean & \C{0.30} & \C{0.26} & \C{0.21} & \C{0.05} & \C{0.23} & \C{0.37} \\
&& Harmonic Mean & \C{0.30} & \C{0.26} & \C{0.21} & \C{0.05} & \C{0.23} & \C{0.37} \\
&& Min NLL & \C{0.30} & \C{0.26} & \C{0.21} & \C{0.05} & \C{0.24} & \C{0.38} \\
&& Max NLL & \C{0.26} & \C{0.27} & \C{0.23} & \C{0.05} & \C{0.24} & \C{0.42} \\
\midrule
\multirow{8}{*}{Boundary} 
& \multirow{4}{*}{Bigram} 
& Arithmetic Mean & {\C{0.50}} & \C{0.29} & \C{0.24} & {\C{0.09}} & \C{0.26} & \C{0.43} \\
&& Harmonic Mean & \C{0.44} & \C{0.30} & \C{0.25} & \C{0.08} & \C{0.27} & \C{0.45} \\
&& Min NLL & \C{0.41} & \C{0.25} & \C{0.18} & \C{0.07} & \C{0.22} & \C{0.33} \\
&& Max NLL & \C{0.36} & \C{0.28} & \C{0.23} & \C{0.06} & \C{0.26} & \C{0.42} \\
\cmidrule(lr){2-9}
& \multirow{4}{*}{Trigram} 
& Arithmetic Mean & \C{0.48} & \C{0.29} & \C{0.26} & \C{0.09} & \C{0.26} & \C{0.47} \\
&& Harmonic Mean & {\C{0.49}} & \C{0.29} & \C{0.26} & \C{0.09} & \C{0.26} & \C{0.47} \\
&& Min NLL & \C{0.44} & \C{0.29} & \C{0.25} & \C{0.08} & \C{0.26} & \C{0.46} \\
&& Max NLL & \C{0.37} & \C{0.30} & \C{0.24} & \C{0.07} & \C{0.27} & \C{0.44} \\
\midrule
\multirow{2}{*}{Baseline} 
& \multicolumn{2}{l}{Uniform Random} & \C{0.15} & \C{0.19} & \C{0.17} & \C{0.03} & \C{0.17} & \C{0.31} \\
& \multicolumn{2}{l}{Stratified Random} & \C{0.17} & \C{0.20} & \C{0.17} & \C{0.03} & \C{0.18} & \C{0.30} \\
\bottomrule
\end{tabular}
\end{table*}

\end{document}